\documentclass[10pt,journal,compsoc]{IEEEtran}

\usepackage{amsmath,amsfonts}
\usepackage{algorithmic}
\usepackage{algorithm}
\usepackage{array}
\usepackage{textcomp}
\usepackage{stfloats}
\usepackage{url}
\usepackage{verbatim}
\usepackage{graphicx}
\usepackage{cite}
\usepackage[hang,small,bf]{caption}
\usepackage[subrefformat=parens]{subcaption}
\usepackage{color,soul} 

\begin{document}

\title{Time-series Anomaly Detection based on Difference Subspace between Signal Subspaces}

\author{
    \IEEEauthorblockN{
        Takumi Kanai\IEEEauthorrefmark{1}, Naoya Sogi\IEEEauthorrefmark{1}, Atsuto Maki\IEEEauthorrefmark{3}, Kazuhiro Fukui\IEEEauthorrefmark{1}
    }
    
    \IEEEauthorblockA{\IEEEauthorrefmark{1} University of Tsukuba}
    \IEEEauthorblockA{\IEEEauthorrefmark{3} KTH Royal Institute of Technology}
}




\IEEEtitleabstractindextext{
\begin{abstract}
This paper proposes a new method for anomaly detection in time-series data by incorporating the concept of difference subspace into the singular spectrum analysis (SSA). The key idea is to monitor slight temporal variations of the difference subspace between two signal subspaces corresponding to the past and present time-series data, as anomaly score. It is a natural generalization of the conventional SSA-based method which measures the minimum angle between the two signal subspaces as the degree of changes. By replacing the minimum angle with the difference subspace, our method boosts the performance while using the SSA-based framework as it can capture the whole structural difference between the two subspaces in its magnitude and direction. We demonstrate our method's effectiveness through performance evaluations on public time-series datasets.
\end{abstract}

}

\maketitle

\section{Introduction}
\IEEEPARstart{T}{his} paper proposes a new method for anomaly detection in time-series data, based on monitoring temporal variation of signal subspace generated using the singular spectrum analysis (SSA).
There are many types of methods for change point detection  \footnote{The principle of our method is based on change point detection. Thus, we use anomaly detection interchangeably with change point detection from time series in this paper. } in time series \cite{aminikhanghahi2017survey,adsurvey,truong2020selective}. They can be roughly divided into two categories: 1) statistics-based methods \cite{ar,kawahara2007change,gustafsson2000adaptive,ref_article5,ide2016change, ref_article6} and 2) deep learning based methods \cite{De_Ryck2021-ch,lstm,gru,8581424}. 

In this paper, we focus on the statistics-based methods that measure the dynamic change between two data distributions corresponding to the past and present time-series data, generated by shifting a sliding window on given time-series data. In general, such a difference can be measured by statistically comparing the probability density functions corresponding to the two distributions. In practice, however, this probability density function is usually approximated by some simple function such as a Gaussian function, since it is often impossible to estimate such a complicated probability density function.

\begin{figure}[tb]
\centering
\includegraphics[scale=0.5]{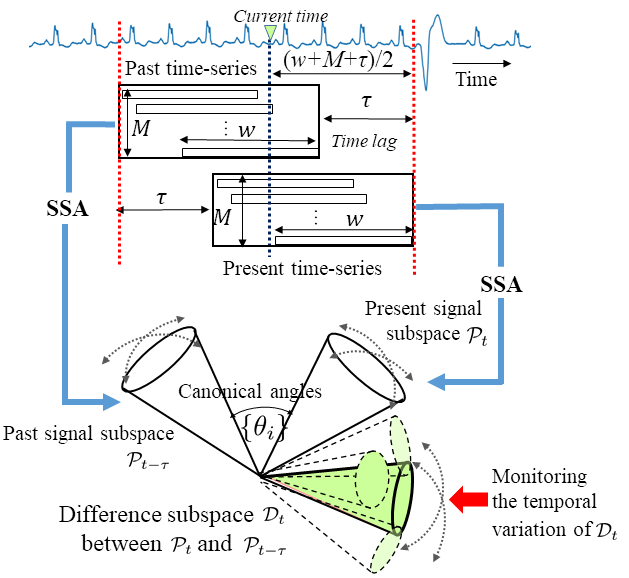}
\caption{The conventional and our methods for anomaly detection. The approaches are different in how to measure the degree of the anomaly change, although both are based on SSA. Conventional method measures the minimum angle $\theta_1$ between the past and present signal subspaces, $\mathcal{P}_{t-\tau}$ and $\mathcal{P}_t$. In contrast, our method measures the temporal variation of the difference subspace $\mathcal{D}_t$ between  $\mathcal{P}_{t-\tau}$ and $\mathcal{P}_t$. } \label{fig:basicidea}
\end{figure}

Among them, in particular, we focus on the mechanism of a method using the singular spectrum analysis (SSA). SSA is a model-free and easy-to-use method for time series analysis \cite{SSAbook}, thus providing a wide range of applications in time-series analysis\cite{mssa,gssa,tssa,mahyub2022}. 
SSA-based method for anomaly detection \cite{ref_article5, ref_article6} relies on a low dimensional subspace, called signal subspace, generated in one of the steps of SSA as shown in Fig.\ref{fig:basicidea}. The main advantage of using signal subspace is that it can represent essential temporal structure of signal data compactly, hence largely reducing the computational cost \cite{ref_article6,mssa,gssa,tssa}. Moreover, the basis of the signal subspace can be stably generated using the singular value decomposition even in the case that learning data is insufficient, unlike the probability function.

The process flow of conventional SSA-based method consists of the following four steps as shown in Fig.\ref{fig:basicidea}. First, the entire time series data is divided into two parts: past and present time series. Second, two signal subspaces are generated by applying the SSA to the past and present time-series data. Next, the minimum angle $\theta_1$ between the present and past subspaces is measured as the degree of anomaly change. Finally, a specific anomaly change is detected when the anomaly score is larger than a given threshold value. Henceforth, we will refer to the above two types of signal subspaces as {\it past} and {\it present} signal subspaces, respectively.

Although the SSA-based method can work well in change detection, there is still a large room for improvement.
The first issue is that the minimum angle between the past and present signal subspaces is often almost zero, producing no information of time change, as the two signal subspaces are generated from two similar, partly overlapping time-series data. The second issue is that the minimum angle cannot capture the whole structural change of the signal subspace even if there is no overlap between the two signal subspaces, since it is just one scalar value. 

\begin{figure}[tb]
\centering
\includegraphics[scale=0.4]{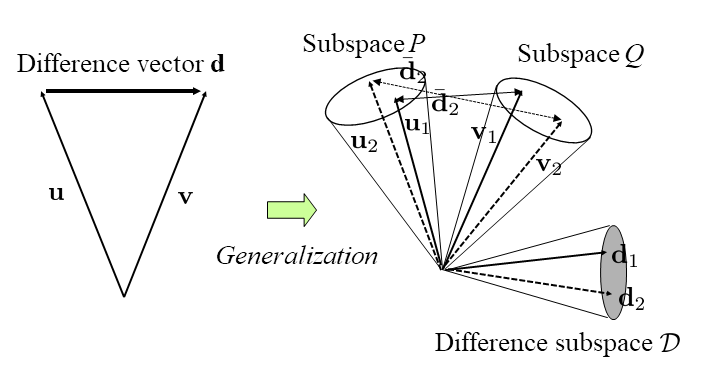}
\caption{Difference subspace $\mathcal{D}$ between subspaces $\mathcal{P}$ and $\mathcal{Q}$.} \label{fig:concept-ds}
\end{figure}

To address these issues and precisely extract the temporal structural change of the signal subspace, we incorporate the concept of difference subspace (DS) \cite{ref_article2}. DS is a natural extension of the difference vector between two vectors, and represents the difference component between two subspaces as shown in Fig.~\ref{fig:concept-ds}. Our basic idea is to detect a subtle temporal variation of the signal subspace through monitoring the temporal structural variation of the DS between the past and present subspaces instead of the minimum angle, as shown in Fig.\ref{fig:basicidea}. 
To the best of our knowledge, this is the first work that applies the general concept of difference subspace (DS) to signal data analysis including anomaly detection, although it has been widely used in tasks of image recognition \cite{ref_article2, tpami2022}.

Here we have one concern: a DS has been originally defined assuming that there is no overlap between two subspaces \cite{ref_article2}. However, to deal with anomaly detection, we need to consider an overlap between two signal subspaces as mentioned earlier. Thus, we revise the definition of DS for the case where there is an overlapping subspace.

To effectively capture the subtle variation of the signal subspace, we define two types of indices regarding the direction and magnitude of the variation. We evaluate the direction by using the canonical angles between an input DS and a reference DS generated from the normal signal data without anomaly change. Here, all the canonical angles are zeros when the directions of two DS's coincide, and they are all 90 degrees when the DS's are orthogonal to each other. We also evaluate the magnitude by the super volume defined according to the geometrical definition of DS. 
Finally, we use the product of the two indices as the degree of anomaly changes in our method. 

Our main contributions are summarized as follows:
\begin{itemize}
    \setlength{\itemsep}{-0.1mm} 
    \setlength{\parskip}{-0.0mm} 
    \item We boost the framework of the SSA-based method for anomaly detection by incorporating the concept of difference subspace between the past and present signal subspaces.
    \item We revise and generalize the definition of difference subspace to deal with the case that there is an overlap between two signal subspaces since we expect this situation in the SSA-based methods.
    \item We introduce a degree of anomaly change considering both the direction and the magnitude of the variation of the signal subspace according to the geometry of difference subspace.
    \item We demonstrate the advantage of our method over the conventional method on public time-series datasets.
\end{itemize}

The rest of the paper is organized as follows. In {Section~\ref{sec:SSA-based method}}, we describe the SSA-based method. In Section ~\ref{sec:proposed method}, we describe the basic idea and framework of the propose method. First, we explain the geometrical definition of difference subspace. Then, we detail the algorithm of our method using difference subspace. In Section ~\ref{sec:evaluation}, we demonstrate the effectiveness of our method through experimental evaluation on several public time-series datasets. Section ~\ref{sec:conclusion} concludes the paper.

\section{SSA-based detection method}\label{sec:SSA-based method}
In this section, 
we first describe how to generate the signal subspace, the core component of the SSA-based method. We then explain the algorithm of the SSA-based method.

\begin{figure}[tb]
\centering
\includegraphics[scale=0.6]{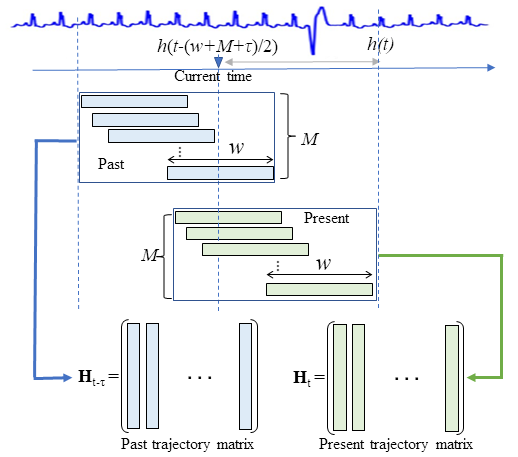}
\caption{Definition of trajectory matrix.}
\label{fig:trajectory}
\end{figure}

\subsection{Generation of signal subspace}
The signal subspace ${\mathcal{P}}_t$ corresponding to a time-series data $h(t)$ is generated by analyzing the trajectory matrix calculated from the time-series data in the process of the SSA as shown in Fig.~\ref{fig:trajectory}. Given one-dimensional time series data $h(t)$, the corresponding trajectory matrix ${\mathbf{H}}_t \in \mathbb{R}^{w{\times}M}$ is defined as follows:

\begin{equation}
\label{eq:hankel}
   {\mathbf{H}}_t=\left[\begin{array}{cccc}
 h(t-w-M+2) & \cdots & h(t-w+1) \\
\vdots & \ddots & \vdots \\
 h(t-M+1) & \cdots & h(t)
 \end{array}\right],
\end{equation}
where $w$ is the width of a sliding window and \textit{M} is the number of the sliding windows as shown in Fig.~\ref{fig:basicidea}. 
To obtain the principal components of $h(t)$, SSA solves the following eigenvalue problem:
  \begin{equation}
   {\mathbf{H}}_{t} {\mathbf{H}}_{t}^{\top} {\mathbf{\Phi}}={\mathbf{\Phi \Sigma}},
   \label{eq:ssa}
  \end{equation}
where $\mathbf{\Phi}$ is the matrix arranging eigenvectors in columns and $\mathbf\Sigma$ is the matrix containing eigenvalues, $\lambda_1, \dots, \lambda_w$, in the diagonal elements.
 The $r$-dimensional signal subspace ${\mathcal{P}}_{t}$ of the input time series data $h(t)$ is spanned by the eigenvectors $\{{\boldsymbol{\phi}}_i\}_{i=1}^r$ corresponding to the $r$ largest eigenvalues in $w$-dimensional vector space.

\subsection{Procedure of anomaly detection}\label{sec:process}
SSA-based methods detect anomaly change in the following procedure.
\begin{enumerate}
\setlength{\itemsep}{-0.1mm} 
\setlength{\parskip}{-0.0mm} 
\item A local time-series data is divided into two overlapping partial data with time lag $\tau$, the past and present time series, as shown in Fig.~\ref{fig:basicidea}. 
\item Trajectory matrices, $H_{t-\tau}$ and $H_{t}$, are calculated from the two partial data, respectively. 
\item Two $r$-dimensional signal subspaces, $\mathcal{P}_{t-\tau}$ and $\mathcal{P}_{t}$, in $w$-dimensional vector space, are generated by using Eq.(\ref{eq:ssa}), from $H_{t-\tau}$ and $H_{t}$, respectively
\item The minimum canonical angle, $\theta_1$, between the two subspaces is calculated. The canonical angles \cite{ref_article4,ref_article7} are calculated by the following singular value decomposition, ${\mathbf{\Phi}}_{r}^{\top} {\mathbf{\Psi}}_{r} = {\mathbf{U \Sigma}} 
{\mathbf{V}}^{\top}$, where ${\mathbf{\Phi}}_r = [{\boldsymbol{\phi}}_1, \dots, {\boldsymbol{\phi}}_r]$ and ${\mathbf{\Psi}}_r = [ {\boldsymbol{\psi}}_1, \dots, {\boldsymbol{\psi}}_r]$ are the basis vectors of the two subspace $\mathcal{P}_{t-\tau}$ and $\mathcal{P}_{t}$, and ${\mathbf{\Sigma}}$ = diag($ [\lambda_1, \dots, \lambda_d] $). $\lambda_i$ corresponds to cos $\theta_{i}$.
\item The dissimilarity between $\mathcal{P}_{t-\tau}$ and $\mathcal{P}_{t}$ is calculated by using the cosine of the smallest canonical angle $\theta_1$ as $1-\lambda_1$. In the following, this value is regarded as the change degree $a(t-t_c)$ at the time, $t-t_c$, for SSA-based methods, where $t_c$ is set to $round((w+M+\tau)/2)$ as the middle point of the time interval from the first of the past to the end of the present. If $a(t-{t_c})$ is higher than a threshold, the time, $t-{t_c}$, is identified as an anomaly change.
\end{enumerate}
The SSA-based methods repeat the above detection process while shifting the sliding window of the input time-series data.

\section{Proposed method}\label{sec:proposed method}
In this section, we first explain the basic idea behind our proposed method. We then describe the definition of the difference subspace (DS), assuming that there is no overlap between the two subspaces. After that, we revise the definition to deal with our case that there can be an overlap between the two signal subspaces in a high dimensional vector space. Finally, we show how to monitor the variation of the difference subspace and the algorithm of our proposed method.

\subsection{Basic idea}
The essence of our method is to monitor the temporal variation of DS between the past and present signal subspaces in a high dimensional vector space. To this end, we consider the variation of DS in two terms: the direction and magnitude. For the direction, we measure the dissimilarity of the present difference subspace $ \mathcal{D}_{in}$ with non-anomalous difference subspaces ${\mathcal{D}_{N}}$ as an index, where the non-anomalous difference subspace is generated from normal time-series data without any change in the learning phase. In this paper, we generate the non-anomalous difference subspace from early time-series data, assuming that there is no anomaly during that term. For the magnitude, we use the sum of the cosines of multiple canonical angles between the past and present signal subspaces as an index. Finally, we use the product of the two indices  as the change degree of our method.

\subsection{Generalization of DS's definition for the SSA-based method}
{\noindent{\bf{Original definition of DS}}}:~We describe the original definition of difference subspace $\mathcal{D}$ \cite{ref_article2} assuming that there is no overlap between two signal subspaces in high dimensional vector space. A DS can be defined in two different ways. In the first one, a DS is geometrically defined as a natural extension of a difference vector $\bar{d}$ between two vectors $u$ and $v$ as shown in Fig~\ref{fig:concept-ds}. 
Given $N_P$-dimensional subspace $\mathcal{P}$ and $N_Q$-dimensional subspace $\mathcal{Q}$ in $w$-dimensional vector space, $N_P$ canonical angles $\lbrace \theta_i \rbrace{^{N_P}_{i=1}}$ (for convenience $N_P\leq N_Q$) can be obtained between them~\cite{ref_article4,ref_article3}, where $\theta_1$ is the minimum angle used in the conventional SSA-based method.

Let ${\bar{\mathbf{d}}}_i \in {\mathbb{R}}^{w}$ be the difference vector, ${\mathbf{v}}_i-{\mathbf{u}}_i$, between canonical vector ${\mathbf{u}}_i$ and ${\mathbf{v}}_i$, which form the $i$th canonical angle, $\theta_i$. As all ${\mathbf{d}}_i$ are orthogonal to each other, the normalized difference vectors ${\mathbf{d}}_i = \frac{{\bar{\mathbf{d}}}_i}{\|{\bar{\mathbf{d}}}_i\|}$ can be regarded as orthonormal basis vectors of the difference subspace $\mathcal{D}$. 

In the second one, DS can be analytically defined by using the orthogonal projection matrices of the two subspaces \cite{ref_article2}. The basis of the difference subspace $\mathcal{D}$ between $\mathcal{P}$ and $\mathcal{Q}$ can be calculated from their projection matrices, $\mathbf{P}$ and $\mathbf{Q}$, which are defined by $\mathbf{P}=\sum_{i=1}^{N_P} \boldsymbol{\phi}_{i} \boldsymbol{\phi}_{i}^{\top} \in {\mathbb{R}}^{w{\times}w}$ and $\mathbf{Q}=\sum_{i=1}^{N_Q} \boldsymbol{\psi}_{i} \boldsymbol{\psi}_{i}^{\top} \in {\mathbb{R}}^{w{\times}w}$,
where $\boldsymbol{\phi}_i \in {\mathbb{R}}^{w}$ and $\boldsymbol{\psi}_i \in {\mathbb{R}}^{w}$ are orthogonal basis vectors of the subspaces $\mathcal{P}$ and $\mathcal{Q}$, respectively. Then, basis vectors of the difference subspace between $\mathcal{P}$ and $\mathcal{Q}$ are calculated from the sum of the projection matrix $\mathbf{G}=\mathbf{P}+\mathbf{Q}$ as follows:

\begin{equation}
 {\mathbf{GD}}={\mathbf{D\Sigma}},
 \label{eq:ds}
\end{equation}
where $\mathbf{D}$ is the matrix arranging eigenvectors $\{{\mathbf{d}}_i\}$ in columns and $\mathbf{\Sigma}$ is the diagonal matrix containing eigenvalues, $\lambda_1, \dots, \lambda_{N_P}$, in the diagonal elements. Difference subspace $\mathcal{D}$ between $\mathcal{P}$ and $\mathcal{Q}$ is spanned by the $N_P$ eigenvectors $\{{d}_i\}$ corresponding to the eigenvalues smaller than one. Finally, the difference subspace $\mathcal{D}$ is defined as Span $( [{\mathbf{d}}_1, {\mathbf{d}}_2, \dots, {\mathbf{d}}_{N_P} ])$. 

\vspace{2mm}
{\noindent\bf{Generalized definition of DS}}:~
In this paper, we need to consider an overlap between two signal subspaces, because such a situation is common in the SSA-based method as mentioned previously. To address this, we generalize the definition of DS assuming that there is an $R$-dimensional overlapping subspace between $\mathcal{P}$ and $\mathcal{Q}$ in $w$-dimensional vector space. In this case, $R$ canonical angles corresponding to the overlap are zero, thus, $\|{{\mathbf{v}}_i-{\mathbf{u}}_i}\|=0$. 
This means that we cannot define the  corresponding $R$ basis vectors, $\{{\mathbf{d}}_i\}$. Accordingly, we revise the definition of DS as $\mathcal{D}$= Span $( [{\mathbf{d}}_1, {\mathbf{d}}_2, \dots, {\mathbf{d}}_{N_P-R} ])$.
From the view of the analytical definition, we also revise the definition as follows: $\mathcal{D}$ is spanned by the $(N_P-R)$ eigenvectors $\{{\mathbf{d}}_i\}$ corresponding to the positive eigenvalues smaller than one of $\mathbf{G}$. 
The $R$ dimensional overlapping subspace is spanned by the eigenvectors corresponding to the eigenvalue two. 
 It is worthwhile to note that $\lambda_i$ is equal to ${\|{\bar{\mathbf{d}}}_i\|}^2$. Therefore, in practice, we use only those eigenvectors corresponding to the eigenvalues smaller than one and larger than a given small value $\delta(>0$, 1e-3 $\sim$ 1e-6) because the eigenvectors $\{{\mathbf{d}}_i\}$ with extremely small eigenvalues are unstable.

\subsection{Measuring the variation of signal subspace}
We describe how to measure the direction and magnitude of the variation of the signal subspace.

\vspace{2mm}
{\noindent\bf{Direction of variation}}:~
We measure the direction of the variation by the dissimilarity of present difference subspace $\mathcal{D}_{in}$ with non-anomalous difference subspace $\mathcal{D}_{N}$, which is generated from normal sequence as will be described in the training phase of the next section. The dissimilarity can be obtained in the same way as described in Sec.\ref{sec:process}. 
Let the basis vectors of $m$-dimensional difference subspace $\mathcal{D}_{in}$ and $n$-dimensional difference subspace $\mathcal{D}_{N}$ be $\mathcal{D}_{in}$ = Span $([{\mathbf{d}}^{in}_1, \dots, {\mathbf{d}}^{in}_m])$ and $\mathcal{D}_{N}$ = Span $([\mathbf{d}^{N}_1, \dots, \mathbf{d}^{N}_n])$, respectively. The canonical angles between the two difference subspaces are calculated as ${\mathbf{D}}_{N}^{\top} {\mathbf{D}}_{in} = {\mathbf{U}} {\mathbf{\Sigma}} {\mathbf{V}}^{\top}$, where ${\mathbf{\Sigma}}$ = diag($ [\lambda_1, \dots,  \lambda_d ]$) and $\lambda_i$ is $\cos \theta{_i}$ between them. With the $c$ smallest canonical angles, the dissimilarity  between them, $\delta$, is calculated 
as
\begin{equation}
\delta(\mathcal{D}_{in}, \mathcal{D}_N) = \frac{1}{c} \sum_{i=1}^{c} (1-\lambda_i). \label{eq:dsim}
\end{equation}
The higher the dissimilarity, the more likely the present signal is anomalous.

\vspace{2mm}
{\noindent\bf{Magnitude of variation}}:~
We measure the magnitude of variation with the super volume defined by using the difference vector as the edge. $\cos \theta_{i}$ of the canonical angles between the past and present signal subspaces corresponds to the length of the difference vector between canonical vectors. As the directions of all the difference vectors are orthogonal, we can calculate the super volumes as the total product of $\{\cos \theta_{i}\}$. We are interested in only the magnitude relation of the super volumes. Thus, we use the logarithmic sum of $\cos \theta_{i}$ as follows:
 \begin{equation}
\mu(\mathcal{D}) = \log (\prod_{i=1}^{d} \cos \theta_i) = \sum_{i=1}^{d} \log \cos \theta_i.
\label{eq:mag1}
\end{equation}
 Let the basis vectors of the past signal subspace $\mathcal{P}_{t-\tau}$ and the present signal subspace $\mathcal{P}_{t}$ be ${\mathbf{\Phi}}_r = [{\boldsymbol{\phi}}_1, \dots, {\boldsymbol{\phi}}_r]$ and ${\mathbf{\Psi}}_r = [ {\boldsymbol{\psi}}_1, \dots, {\boldsymbol{\psi}}_r],$ respectively. The cosine of the canonical angles between them are calculated by solving the following singular value decomposition, $ {\mathbf{\Phi}}_{r}^{\top} {\mathbf{\Psi}}_{r} = {\mathbf{U}} {\mathbf{\Sigma}} {\mathbf{V}}^{\top}$, where $\mathbf\Sigma$ = diag($ [\cos \theta_1, \dots, \cos \theta_d] $).
 
 Let $\mu(\mathcal{D}_{N})$ be the average of $L$ variations, $\{\mu(\mathcal{D}_{i})\}_{i=1}^L$, obtained from a normal sequence in the training phase. Finally, we calculate the variation of magnitude of an input difference subspace, $\mathcal{D}_{in}$, by $\beta=({\mu}(\mathcal{D}_{in}) - {\mu}(\mathcal{D}_{N}))^2$. 

\vspace{2mm}
{\noindent\bf{Definition of change degree}}:~
To take into account the two indices  at the same time, we consider the change degree $\hat{a}(t)$ defined by the product of 
$\beta$ and $\delta(\mathcal{D}_{in}, \mathcal{D}_N)$ as 
\begin{equation}
\hat{a}(t) =  \beta {\times}\delta(\mathcal{D}_{in}, \mathcal{D}_N).\label{eq:mag2}
\end{equation}
The larger ${\hat{a}}(t)$, the more likely the present signal is anomalous.

\begin{figure}[tb]
\centering
\includegraphics[scale=0.4]{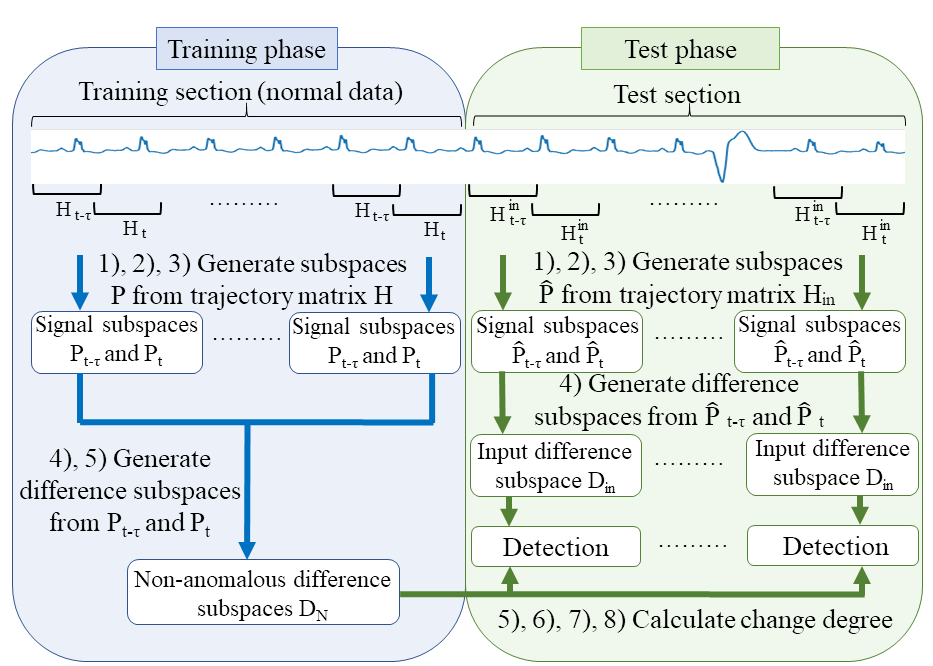}
\caption{Process flow of the proposed method.}
\label{fig:flow}
\end{figure}

\subsection{Process flow of the proposed method}
In this section, we describe the whole process flow of the proposed method from generating two signal subspaces to monitoring the difference subspace and further detecting anomaly changes.
The proposed method consists of two phases, the training phase and the detection phase as shown in Fig.~\ref{fig:flow}.

\vspace{2mm}
{\noindent\bf{Training phase}}:~
\begin{enumerate}
\setlength{\itemsep}{-0.1mm} 
\setlength{\parskip}{-0.0mm} 
 \item  A normal time-series data is divided into present $h(t)$ and past $h(t-\tau)$ shifted by $\tau$. 
 \item The trajectory matrices $H_{t}$ and $H_{t-\tau}$ are calculated from the two time series data $h(t)$ and $h(t-\tau)$ as shown in Eq.(\ref{eq:hankel}).
 \item The basis vectors, ${\mathbf{\Phi}} = [{\boldsymbol{\phi}}_1, \dots, {\boldsymbol{\phi}}_r]$ and ${\mathbf{\Psi}} = [{\boldsymbol{\psi}}_1, \dots, {\boldsymbol{\psi}}_r]$, of the past and present subspaces, $\mathcal{P}_{t-\tau}$ and $\mathcal{P}_{t}$, are obtained from the trajectory matrices by Eq.(\ref{eq:ssa}).
 \item The basis vectors $\mathcal{D}=[{d}_1, \dots, {d}_n]$ of the difference subspace $\mathcal{D}$ between $\mathcal{P}_{t-\tau}$ and $\mathcal{P}_{t}$ are calculated by using Eq.(\ref{eq:ds}).
 \item The magnitude of variation ${\mu}(\mathcal{D})$ between $\mathcal{P}_{t-\tau}$ and $\mathcal{P}_{t}$ are calculated by using Eq.(\ref{eq:mag1}).
\end{enumerate}
We repeat the process from 1 to 5 over the normal time-series data to obtain a set of $L$ non-anomalous difference subspaces $\lbrace \mathcal{D}_{i} \rbrace{^L_{i=1}}$ and their normal magnitudes, $\{{\mu}(\mathcal{D}_{i})\}_{i=1}^L$.
Finally, we define non-anomalous difference subspace, $\mathcal{D}_{N}$, as the principal component subspace of $L$ non-anomalous difference subspaces which is spanned by the $nor\_dims$ eigenvectors of $\sum_{i=1}^{L}\mathcal{D}_i\mathcal{D}_i^{\top}$. $nor\_dims$ is the dimension of the non-anomalous difference subspace. Furthermore, we calculate the average magnitude, ${\mu}(\mathcal{D}_{N})$, as $\frac{1}{L}\sum_{i=1}^L {\mu}(\mathcal{D}_{i})$.

\vspace{2mm}
{\noindent\bf{Detection phase}}:~
\begin{enumerate}
\setlength{\itemsep}{-0.1mm} 
\setlength{\parskip}{-0.0mm} 
 \item  An input time-series data is divided into present $h_{in}(t)$ and past $h_{in}(t-\tau)$ shifted by $\tau$. 
 \item The trajectory matrices $H^{in}_{t}$ and $H^{in}_{t-\tau}$ are calculated from the time-series data $h_{in}(t)$ and $h_{in}(t-\tau)$ in the same way as in the training phase.
 \item The basis vectors $\Phi$ and $\Psi$ of subspace of $\mathcal{\hat{P}}_{t-\tau}$ and $\mathcal{\hat{P}}_{t}$, respectively, are each generated from $H^{in}_{t}$ and $H^{in}_{t-\tau}$ by using Eq.(\ref{eq:ssa}).
 \item The basis vectors of difference subspace $\mathcal{D}_{in}$ between $\mathcal{\hat{P}}_{t-\tau}$ and $\mathcal{\hat{P}}_t$ are generated in the same way as in the training phase. Let the basis vectors of the difference subspace $\mathcal{D}_{in}$ be $[{\mathbf{d}}_1, ... , {\mathbf{d}}_m]$.
 \item Dissimilarity $\delta(\mathcal{D}_{in}, \mathcal{D}_N)$ between the input difference subspace, $\mathcal{D}_{in}$, and the non-anomalous difference subspace, $\mathcal{D}_N$, is generated in the training phase.
 \item The magnitude index $\beta=({\mu}(\mathcal{D}_{in}) - {\mu}(\mathcal{D}_{N}))^2$ is calculated. 
\item Change degree at the time, $t-{t_c}$, is calculated using Eq.(\ref{eq:mag2}), where $t_c$ is set to $round((w+M+\tau)/2)$ as the middle point of the time interval from the first of the past to the end of the present. Then, this degree is regarded as a final change degree $\hat{a}(t-{t_c})$ at the time, $t-{t_c}$.
\item If the change degree $\hat{a}(t-{t_c})$ is larger than a threshold, the time, $t-{t_c}$, is identified as an anomaly change, where the threshold is obtained as the average of the change degrees of the normal signal subspaces.
\end{enumerate}
We repeat the above detection process while sliding the time period of the input time-series data.

\section{Evaluation experiments}\label{sec:evaluation}
In this section, we demonstrate the effectiveness of our method by comparison experiments with several conventional methods on seven kinds of public datasets.

\subsection{Performace evaluation in terms of AUC}
\noindent{\bf{Datasets}}:~We used seven datasets from the UCR Time Series Data Mining Archive \cite{Eamonn2005-ea}: chfdb\_chf01\_275\_1, chfdb\_chf01\_275\_2, mitdb\_\_100\_180\_1, mitdb\_\_100\_180\_2, nprs44, stdb\_308\_0\_1 and stdb\_308\_0\_2. Note that as we use the first part of a dataset as normal data in the training phase, we selected the above datasets that do not contain any anomalies in their first part. They are electrocardiogram data and respiratory data during sleep with the labels of normal/anomalous of each data. We divided the whole sequence data of each dataset into two data in the ratio of approximately 30\%:70\%. The former was extracted from the first to a certain time and used as training data for generating non-anomalous difference subspaces and learning the models of the conventional methods. The latter was used for testing. 
The lengths of the training and test data are (1000, 2500) for chfdb\_chf01\_275\_1, chfdb\_chf01\_275\_2, mitdb\_\_100\_180\_1 and mitdb\_\_100\_180\_2, (2800, 6500) for nprs44, and (1500, 3500) for stdb\_308\_0\_1 and stdb\_308\_0\_2.

\vspace{2mm}
{\noindent\bf{Comparison methods}}:~
We consider the naive SSA-based method using only the smallest canonical angle as the baseline method, and refer to it to as SSA$\_\theta_1$.     
Besides, we evaluated three conventional methods: AR (Auto regression)-based method \cite{ar}, GRU (Gated recurrent unit) \cite{gru}, and LSTM (Long short term memory) \cite{lstm}. 
The first method belongs to machine learning-based methods with a classical statistical model, and the remaining methods belong to deep neural network-based methods. We compared two different types of change detection methods accordingly. 

The AR-based method is based on the prediction of the present signal by the linear equation using the past signals. It detects anomaly change points by using the difference between the predicted signal and the observed present signal. In this experiment, the length of the past signals for the prediction was set by the Akaike information criterion \cite{ar,aic}. 
The detection flow using GRU and LSTM is the same as that of the AR-based method, i.e. we first predict the present signal by GRU or LSTM-based network, then measure the difference between the prediction and observation.
Note that the principle of AR, GRU, and LSTM is to predict the present signal using a model generated from the past time series, unlike our method and the baseline that directly observe the change in the signal subspace. 
In this experiment, we use a two-layer network for LSTM and GRU. 

\vspace{2mm}
{\noindent\bf{Performance metrics}}:~We used the Area Under the Curve (AUC) to evaluate the performance of the methods. AUC is calculated from change degree data and correct labels.

\begin{figure}[tb]
\centering
\includegraphics[scale=0.45]{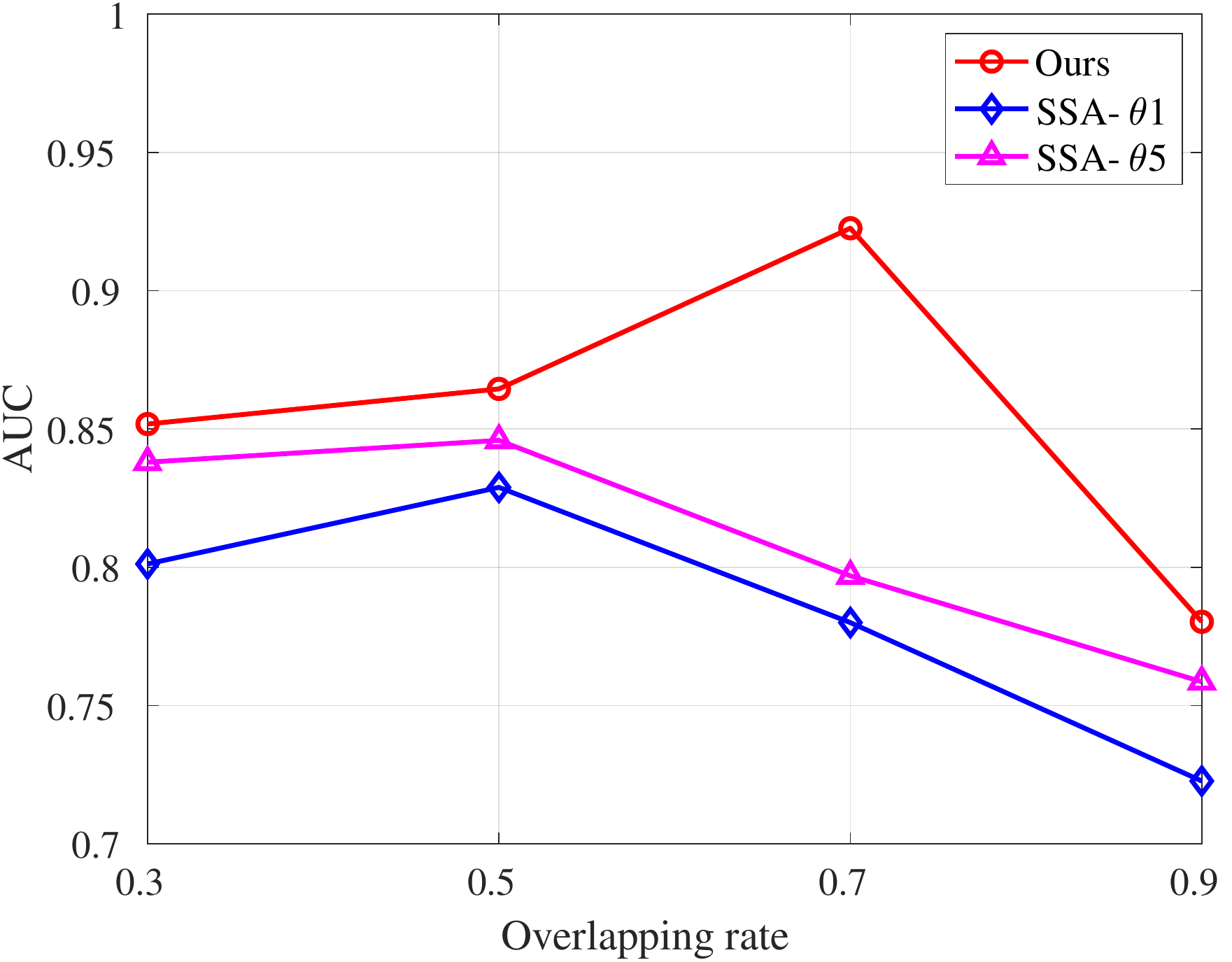}
\caption{AUC vs. overlap rate between the past and present time periods.}
\label{fig:auc}
\end{figure}

\begin{table}[tb]
\centering
\caption{The experimental results of AUC for 10 datasets provided by the UCR time series data mining archive. }\label{tab:auc}
\begin{tabular}{m{23mm}|m{5mm}m{5mm}m{6mm}m{6mm}m{7mm}|l}
\hline
datasets &  AR & GRU & LSTM & SSA$\_\theta_1$ & SSA$\theta_5$ & Ours\\
\hline
chfdb\_chf01\_275\_1 & 0.583 & 0.882 & 0.881 & 0.879 & 0.913 & {\bfseries 0.992}\\
chfdb\_chf01\_275\_2 & 0.474 & 0.681 & 0.937 & 0.972 & 0.970 & {\bfseries 0.977}\\
mitdb\_\_100\_180\_1 & 0.529 & 0.553 & 0.544 & 0.883 & 0.896 & {\bfseries 0.989}\\
mitdb\_\_100\_180\_2 & 0.476 & 0.583 & 0.621 & 0.857 & 0.902 & {\bfseries 0.973}\\
nprs44 & 0.618 & {\bfseries 0.715} & 0.688 & 0.696 & 0.776 & 0.690\\
stdb\_308\_0\_1 & 0.510 & 0.722 & 0.685 & 0.818 & 0.839 & {\bfseries 0.928}\\
stdb\_308\_0\_2 & 0.529 & 0.636 & 0.649 & 0.695 & 0.626 & {\bfseries 0.908}\\
\hline
average & 0.531 & 0.682 & 0.715 & 0.829 & 0.846 & {\bfseries 0.923}\\
\hline
\end{tabular}
\end{table}

 \begin{figure*}[tb]
  \begin{minipage}{0.15\linewidth}
    \centering
    \includegraphics[keepaspectratio, scale=0.14]{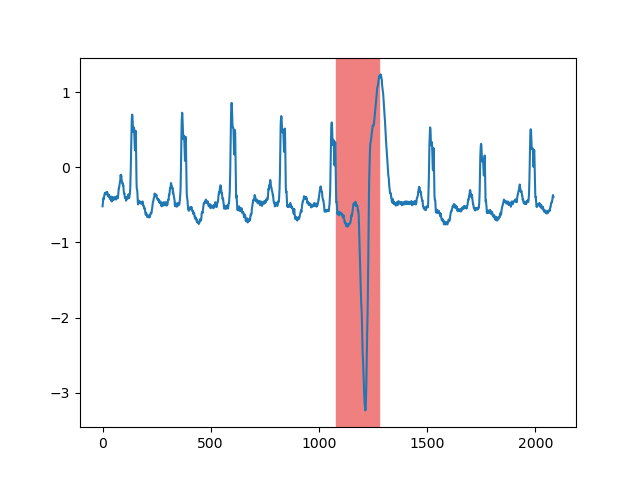}
  \end{minipage}
  \begin{minipage}{0.15\linewidth}
    \centering
    \includegraphics[keepaspectratio, scale=0.14]{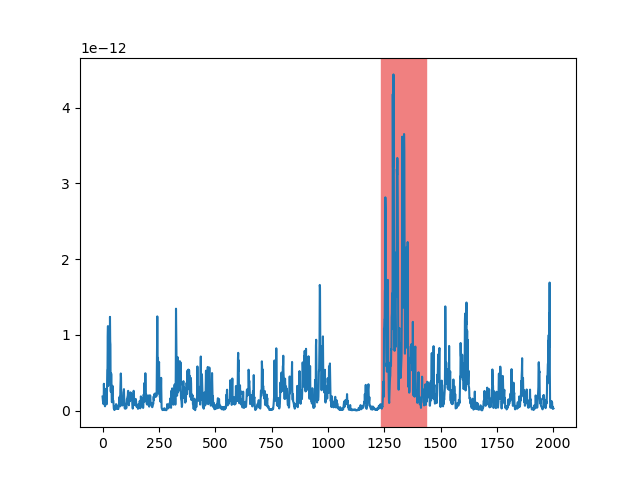}
  \end{minipage}
  \begin{minipage}{0.15\linewidth}
    \centering
    \includegraphics[keepaspectratio, scale=0.14]{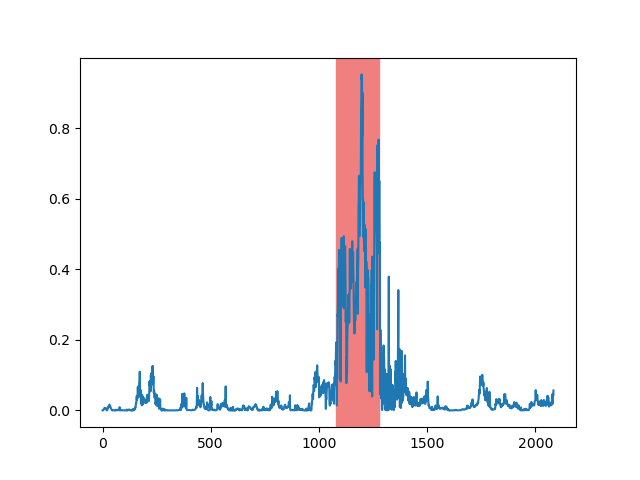}
  \end{minipage}
  \begin{minipage}{0.15\linewidth}
    \centering
    \includegraphics[keepaspectratio, scale=0.14]{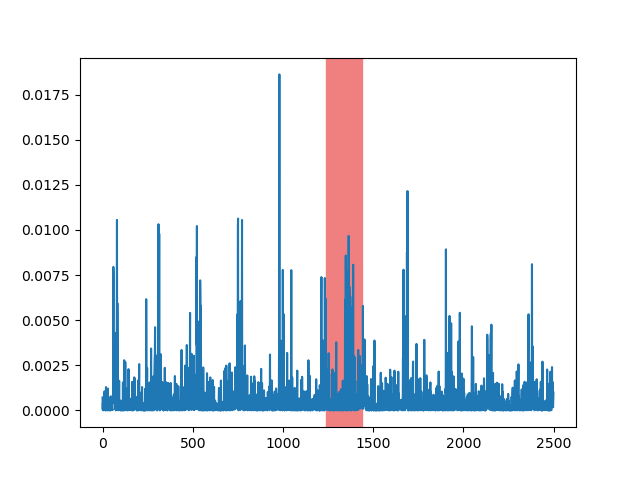}
  \end{minipage}
  \begin{minipage}{0.15\linewidth}
    \centering
    \includegraphics[keepaspectratio, scale=0.14]{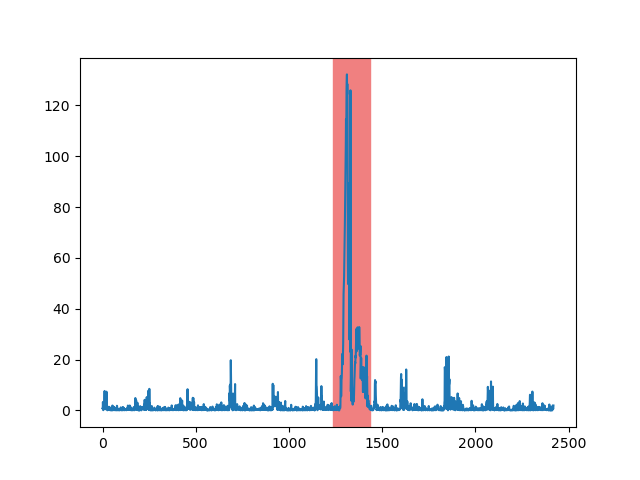}
  \end{minipage}
  \begin{minipage}{0.15\linewidth}
    \centering
    \includegraphics[keepaspectratio, scale=0.14]{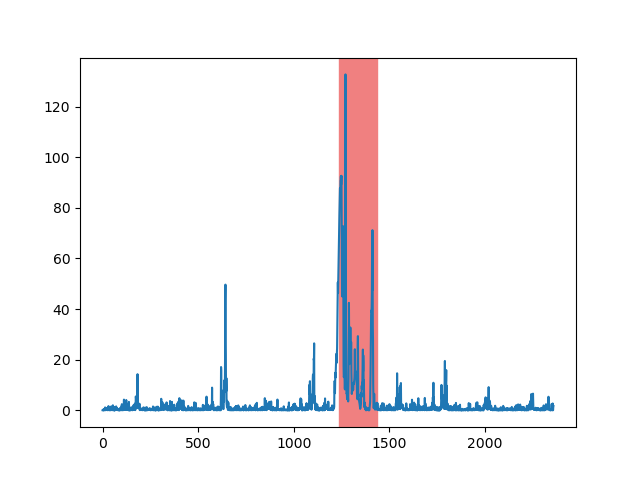}
  \end{minipage}
  \\
  \begin{minipage}{0.15\linewidth}
    \centering
    \includegraphics[keepaspectratio, scale=0.14]{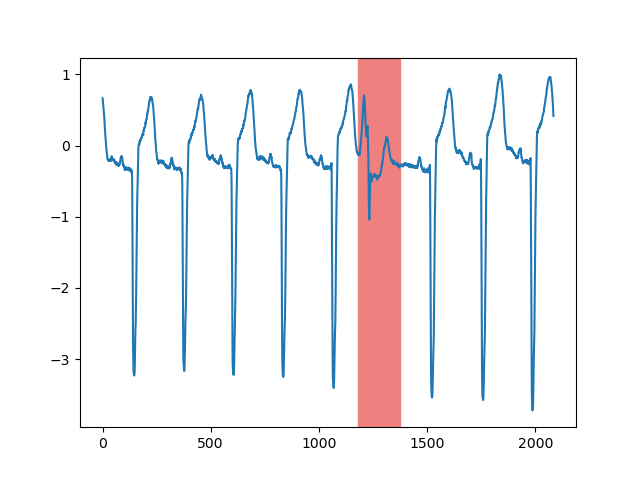}
  \end{minipage}
  \begin{minipage}{0.15\linewidth}
    \centering
    \includegraphics[keepaspectratio, scale=0.14]{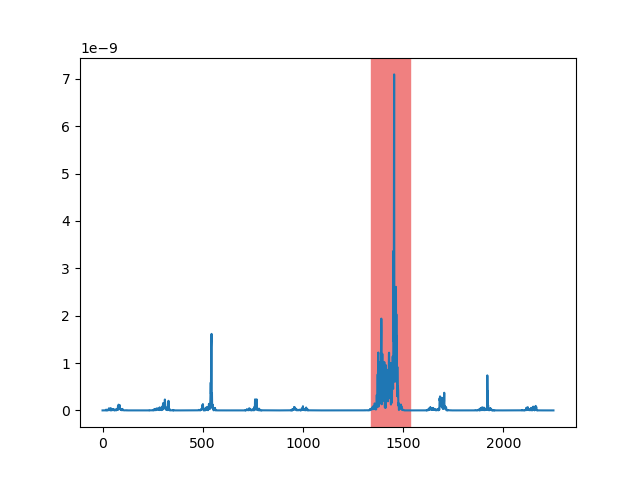}
  \end{minipage}
  \begin{minipage}{0.15\linewidth}
    \centering
    \includegraphics[keepaspectratio, scale=0.14]{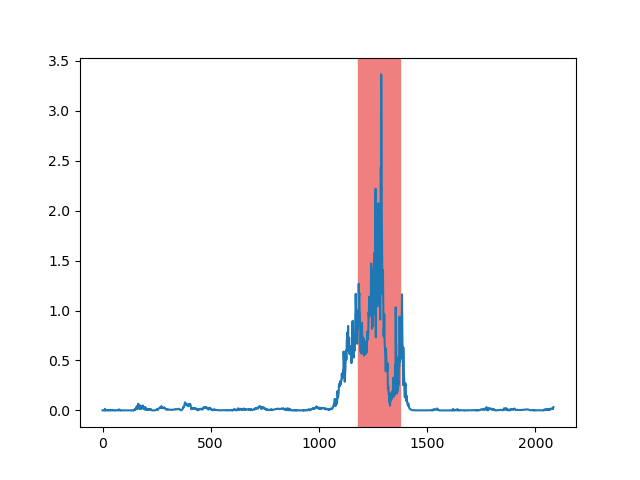}
  \end{minipage}
  \begin{minipage}{0.15\linewidth}
    \centering
    \includegraphics[keepaspectratio, scale=0.14]{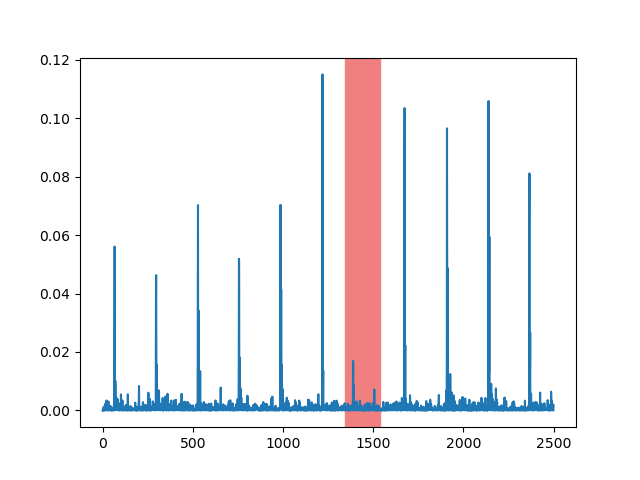}
  \end{minipage}
  \begin{minipage}{0.15\linewidth}
    \centering
    \includegraphics[keepaspectratio, scale=0.14]{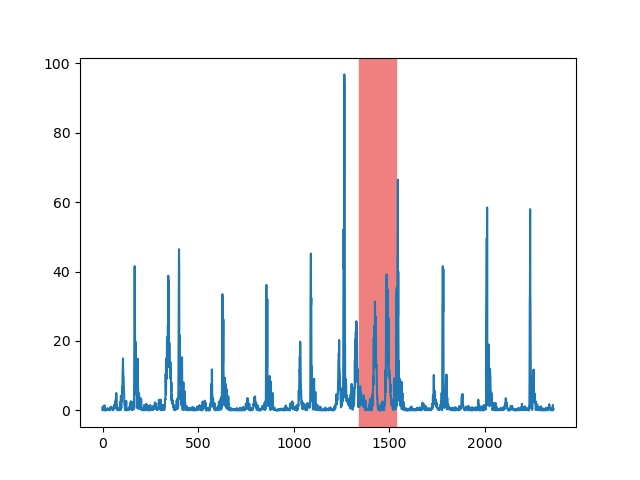}
  \end{minipage}
  \begin{minipage}{0.15\linewidth}
    \centering
    \includegraphics[keepaspectratio, scale=0.14]{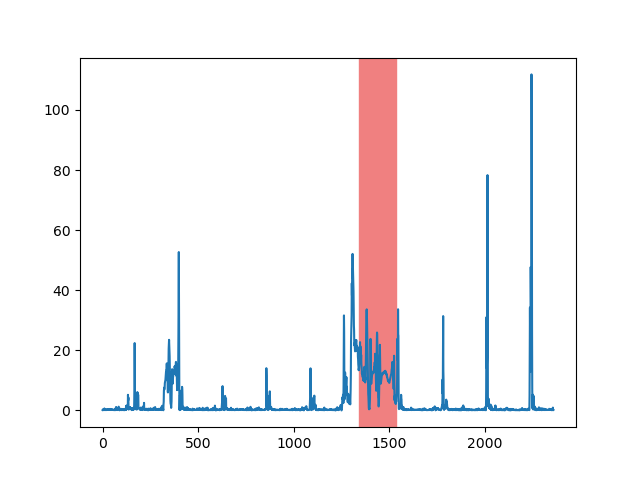}
  \end{minipage}

  \begin{minipage}{0.15\linewidth}
    \centering
    \includegraphics[keepaspectratio, scale=0.14]{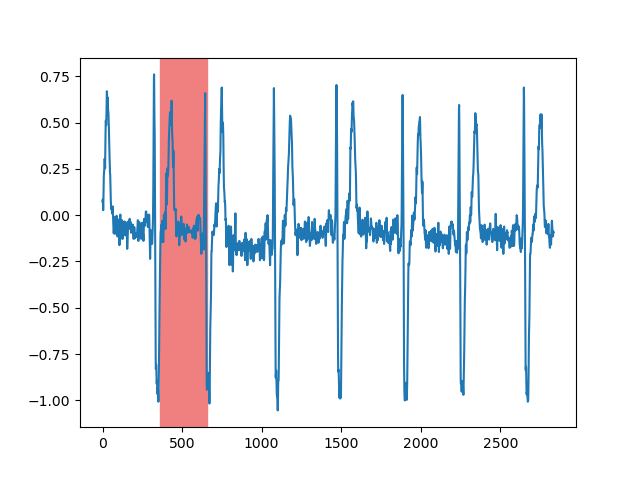}
    \subcaption{Test data}
  \end{minipage}
  \begin{minipage}{0.15\linewidth}
    \centering
    \includegraphics[keepaspectratio, scale=0.14]{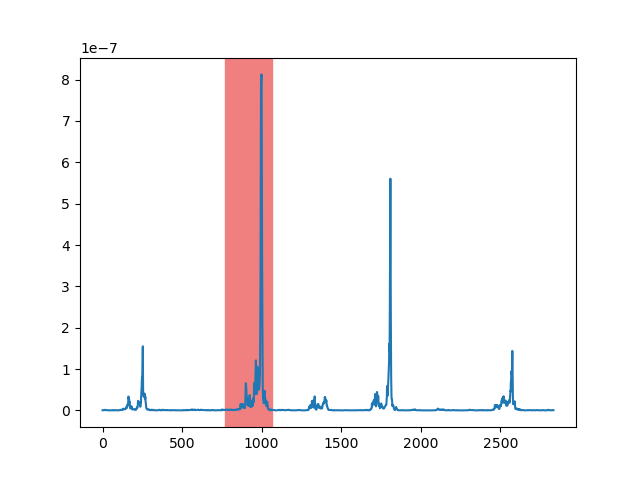}
    \subcaption{SSA$\_\theta_{1}$}
  \end{minipage}
  \begin{minipage}{0.15\linewidth}
    \centering
    \includegraphics[keepaspectratio, scale=0.14]{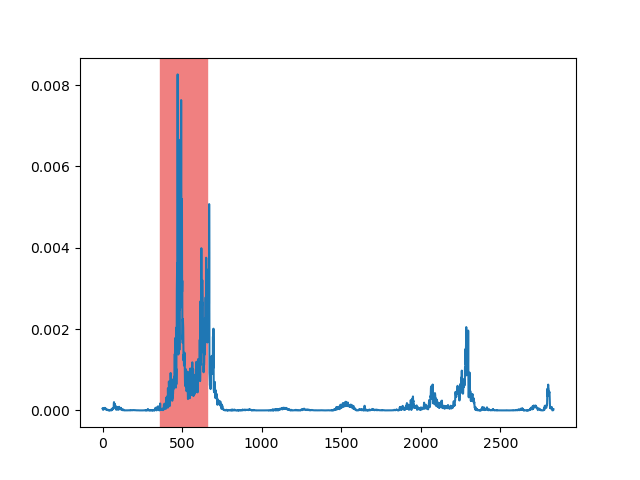}
    \subcaption{Ours}
  \end{minipage}
  \begin{minipage}{0.15\linewidth}
    \centering
    \includegraphics[keepaspectratio, scale=0.14]{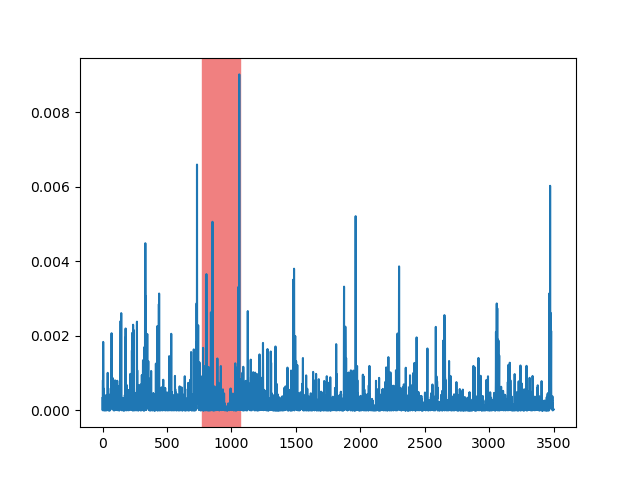}
    \subcaption{AR}
  \end{minipage}
  \begin{minipage}{0.15\linewidth}
    \centering
    \includegraphics[keepaspectratio, scale=0.14]{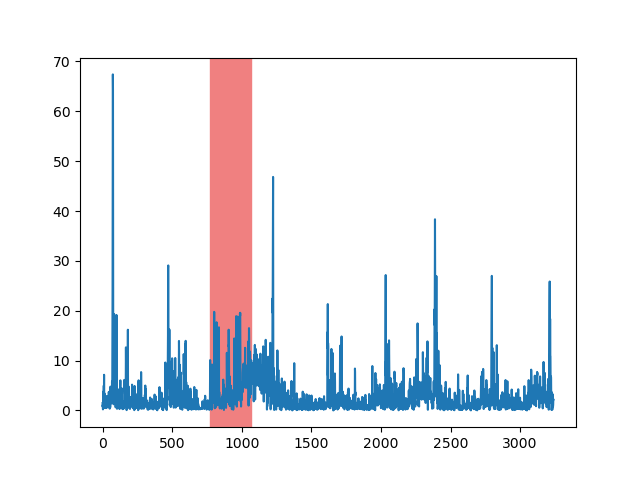}
    \subcaption{GRU}
  \end{minipage}
  \begin{minipage}{0.15\linewidth}
    \centering
    \includegraphics[keepaspectratio, scale=0.14]{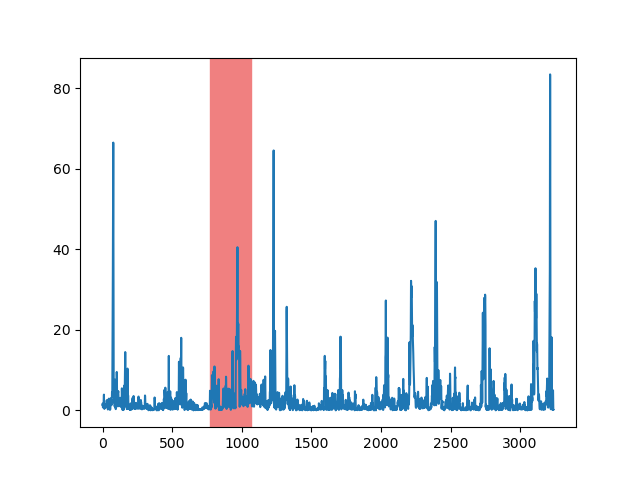}
    \subcaption{LSTM}
  \end{minipage}
  \caption{Change degrees by the proposed method, SSA$\_\theta_{1}$ and SSA$\_\theta_{5}$ for chfdb\_chf01\_275\_1, chfdb\_chf01\_275\_2 and stdb\_308\_0\_1 from the top. Test data contains anomalies as shown in the red range.}
  \label{fig:change}
\end{figure*}

\vspace{2mm}
{\noindent\bf{Parameters}}:~
The parameters of LSTM and GRU are the number of units \textit{u} in each layer and the length \textit{l} of the time series used for prediction. We searched for the best AUC while changing the parameters as $\textit{u} = \lbrace 8, 16, 32 \rbrace$, $\textit{l} = \lbrace 64, 128, 256 \rbrace$. We used the best parameters for each data. 

The parameters of our method are as follows: the width of a sliding window, \textit{w}, the number of the slide windows, \textit{M}, the overlap rate of the past and present sequences \textit{ov\_rate}, the dimension of signal subspace \textit{sig\_dims}, the minimum eigenvalue,$\delta$, used to generate a difference subspace, the dimension of the non-anomalous difference subspace \textit{nor\_dims}, and the number of canonical angles used to calculate dissimilarity \textit{c}. We set the \textit{sig\_dims} to 30  according to the cumulative contribution of 95\% in terms of the eigenvalues. 
The parameters \textit{w}, \textit{M}, and \textit{ov\_rate} are dominant for our method. Thus, the remaining minor parameters are fixed as $\delta$ = 1e-6, $nor\_dims$ = 90$, $c = 5. For the parameters of the baselines, SSA$\_\theta_{1}$ and SSA$\_\theta_{5}$, we used the same as in our method.
The combination of $\textit{w}$ and $\textit{M}$ determine a signal subspace. We searched the best AUC while changing the combination of $\textit{w} = \lbrace 64, 128, 256 \rbrace$, $\textit{M} = \lbrace 64, 128, 256 \rbrace$ in terms of each \textit{ov\_rate} = $\lbrace 0.3, 0.5, 0.7, 0.9 \rbrace$. 


\vspace{2mm}
{\noindent\bf{Results and consideration}}:~
First of all, we consider the advantage of our method over the baselines. Fig.\ref{fig:auc} shows the best AUC for our method and the baselines in terms of the overlap rate \textit{ov\_rate} of the sequences of the past and present.
The AUC of our method is much higher than that of the baseline. This reflects the merit of using the directions and magnitudes of multiple canonical angles instead of only the smallest canonical angle for capturing slight temporal variations of signal subspaces.
Fig.\ref{fig:change} shows examples of anomaly scores of the different methods on all the datasets. From this figure, we can also observe the superiority of the proposed method over SSA$\_\theta_{1}$. 
The proposed method could detect slight change points more clearly and stably while restraining the other normal part without over-detection. On the other hand, the baseline miss-detected several normal change points as anomaly change points.

Next, we compare our method with the conventional methods, AR, GRU, and LSTM. Table~\ref{tab:auc} shows the best AUC for all the methods.
We can see that they do not work as expected. This could be due to the small sample size and data complexity that impairs these methods from learning the valid models of the given data. In contrast, for our method and the baseline, the signal subspace can be generated stably even from a small number of samples, thus resulting in better performance.

\subsection{Sensitivity analysis in terms of signal subspace dimension} \label{sec:evaluation2}
To see the robustness of our method, we compared our method's performances with the baselines, SSA$\_{\theta_1}$ and SSA$\_\theta_{all}$, while changing the dimension of the signal subspace on UCR data used in the previous experiment.
SSA$\_{\theta_1}$ and SSA$\_{\theta_{all}}$ are the methods using first canonical angles and all canonical angles, respectively. 
We used the best parameters found in the previous experiment.
For the dimension of the past and present signal subspaces, we varied it in the range of $ \lbrace 1, 5, 10, ..., 50 \rbrace$.

\begin{figure*}[tb]
  \begin{minipage}[b]{0.5\linewidth}
    \centering
    \includegraphics[keepaspectratio, scale=0.2]{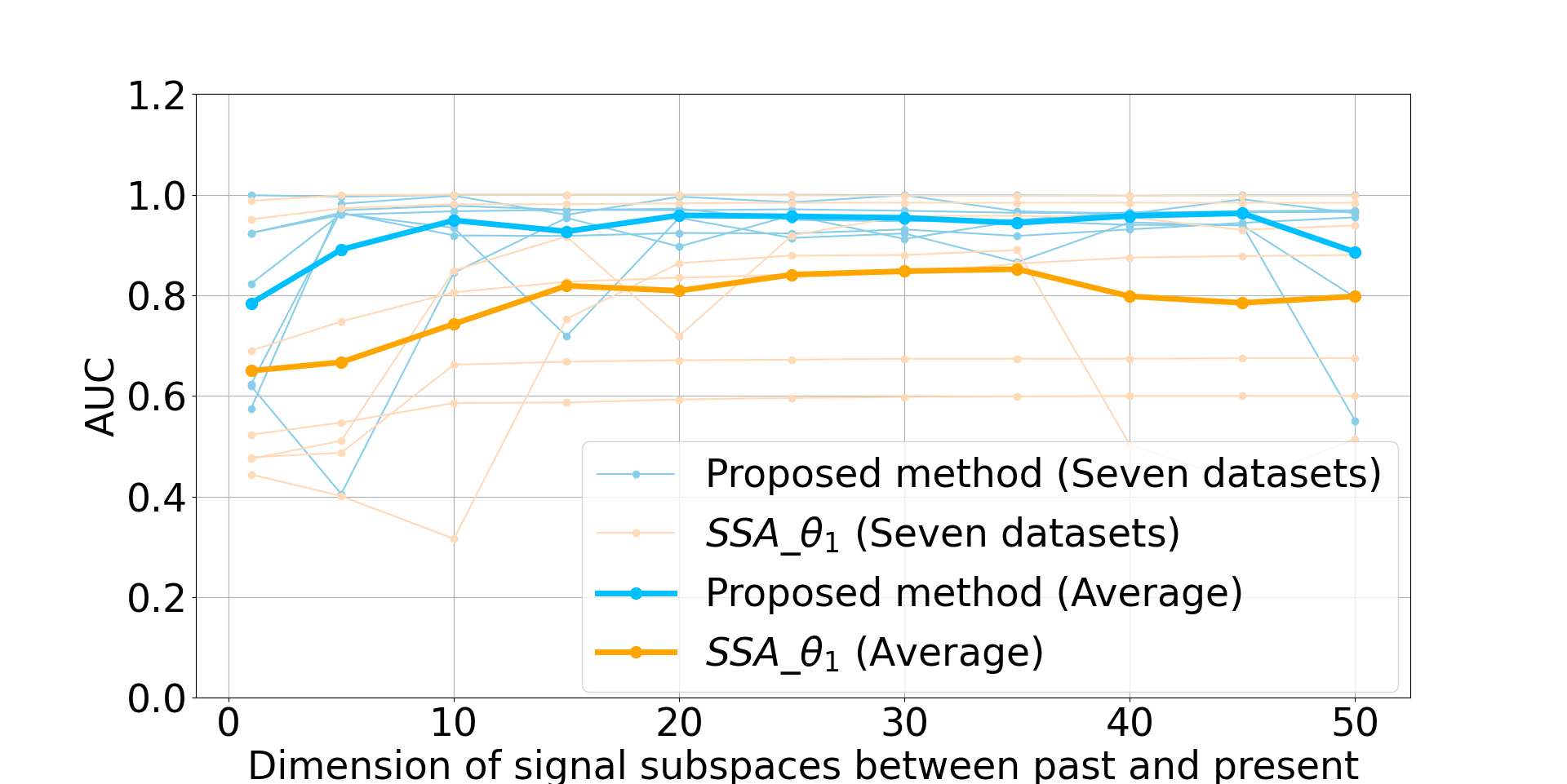}
  \end{minipage}
  \begin{minipage}[b]{0.5\linewidth}
    \centering
    \includegraphics[keepaspectratio, scale=0.2]{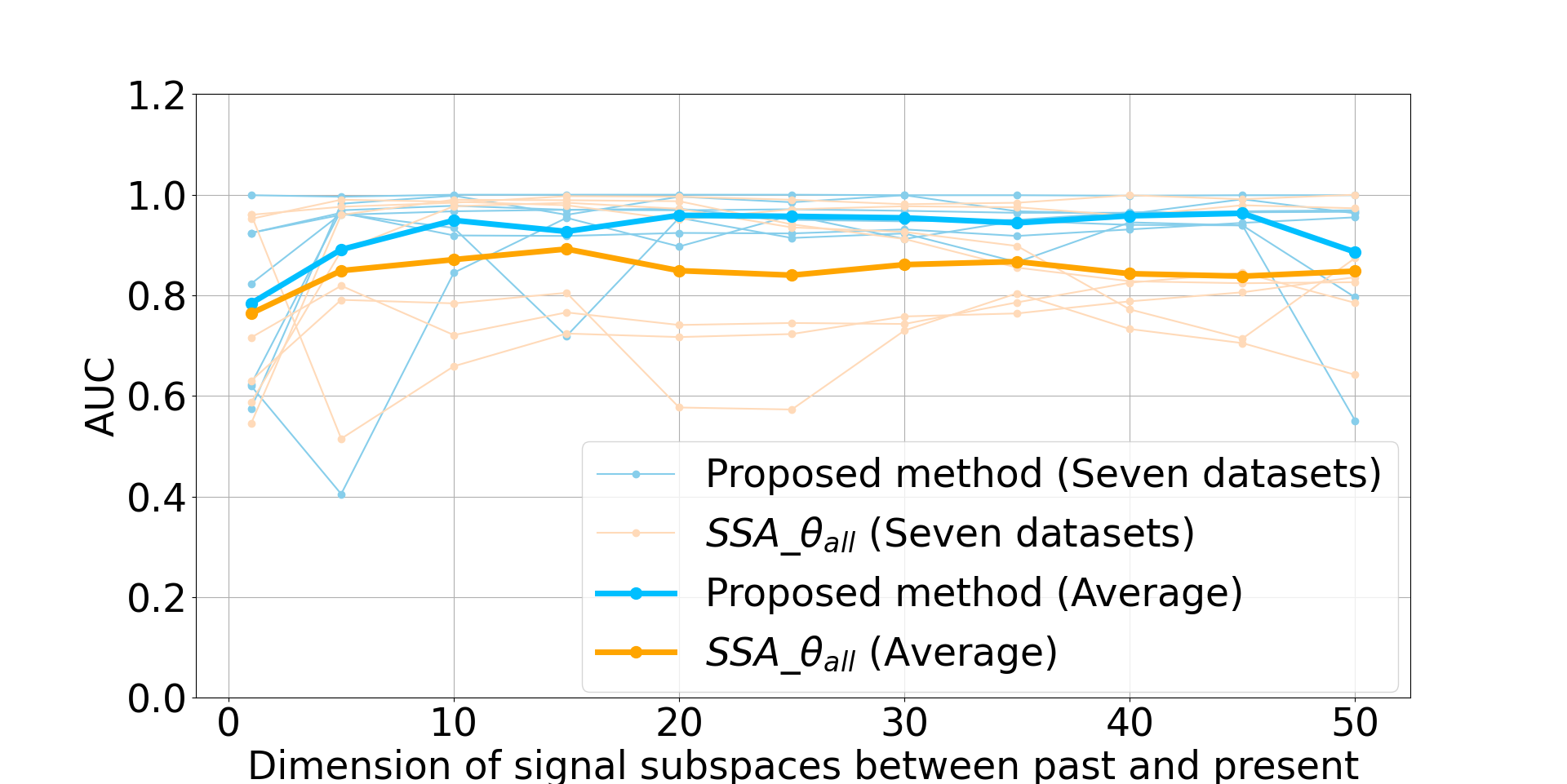}
  \end{minipage}
  \caption{Performances of our method and the baseline method depending on the dimension of signal subspaces.}
  \label{fig:robust}
\end{figure*}

Fig.\ref{fig:robust} shows the change in the performances of both methods across the different dimensions in terms of AUC. We can see that our method achieves a higher AUC than the baseline methods across almost all the dimensions, indicating that the concept of difference subspace contributes to the improvement regardless of the dimension of the signal subspaces. 

\subsection{Visualization of distributions of normal and anomalous time-series data}
In the previous experiment, we showed that our method outperforms the conventional method in terms of AUC. Here, we study the advantage by visualizing how two distributions of normal and anomalous time-series data are separated with each metric of both methods. The conventional method uses the distances (minimum angle) between two signal subspaces as a metric, whereas our method uses the distances between two difference subspaces.

To visualize the distributions of these subspaces in 3D space, we introduce a Grassmann manifold on which each subspace with the same dimension is represented as a point. Then, we apply the Multi-Dimensional Scaling (MDS) using the above metrics. The dimensions of the signal subspace and the difference subspace were set to 30 and 15, respectively. We used the best values obtained in the previous experiment for the other parameters. 

\begin{figure*}[tb]
  \begin{minipage}[b]{0.3\linewidth}
    \centering
    \includegraphics[keepaspectratio, scale=0.4]{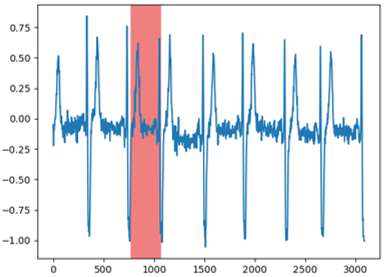}
    \subcaption{Input data \\(stdb\_308\_0\_1)} 
  \end{minipage}
  \begin{minipage}[b]{0.3\linewidth}
    \centering
    \includegraphics[keepaspectratio, scale=0.4]{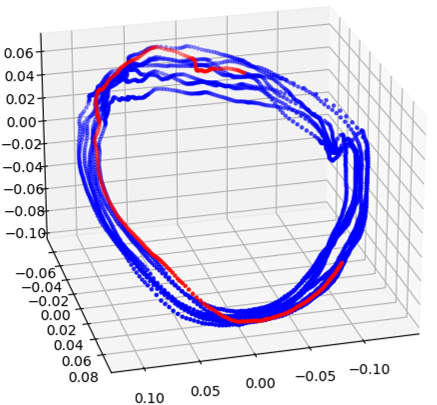}
    \subcaption{SSA$\_\theta_{1}$}
  \end{minipage}
  \begin{minipage}[b]{0.3\linewidth}
    \centering
    \includegraphics[keepaspectratio, scale=0.44]{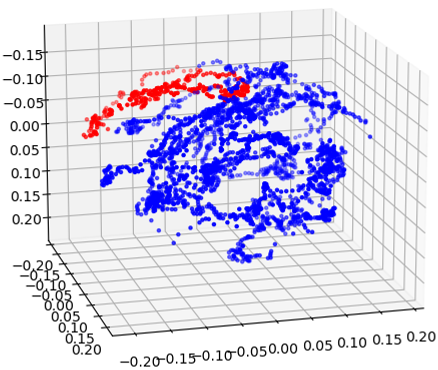}
    \subcaption{Our method}
  \end{minipage}
  \caption{Two distributions of non-anomalous and anomalous subspaces on stdb\_308\_0\_1. Blue points indicate non-anomalous subspaces, and red points indicate anomalous subspaces.}
  \label{fig:visual}
\end{figure*}

Fig. \ref{fig:visual} shows input time-series data (stdb\_308\_0\_1) and the visualizations of two distributions of normal and anomalous subspaces generated by both of the methods. Fig. \ref{fig:visual} (b) and (c) visualize the difference between the conventional method and our methods respectively. The blue points indicate non-anomalous subspaces, and the red points anomalous subspaces.
We can see that the SSA-based method cannot separate the two distributions. In contrast, our method allows the separation. These results support the benfit of our method in class separability.

\section{Conclusion} \label{sec:conclusion}
In this paper, we proposed a method based on the singular spectral analysis (SSA) for detecting anomaly change in time-series data. Our basic idea is to monitor slight temporal variations of signal subspace through the difference subspace between the past and present signal subspaces which we generate by applying the SSA to the past and present time-series data, respectively. This work is the first application of difference subspace, an extension of the difference vector between two vectors, to anomaly detection.
We compared our method with the baseline, the conventional SSA-based method using the minimum angle between the two signal subspaces, on seven public datasets while testing various signal subspace dimensions. Furthermore, we compared our method with some conventional methods, such as AR-based method, GRU-based method and LSTM. We demonstrated the effectiveness of our method through the basic but essential evaluation experiments and visualizing the distributions of normal and anomalous time-series data.

The performance of our method relies on the main parameters: the time lag $\tau$, the size of a sliding window $w$, and the number of sliding windows $M$. The first interesting future work is to develop a method for effectively tuning the parameters by considering the geometrical characteristics of signal and difference subspace. The second extension will be to consider online learning for generating and updating a normal signal subspace from the detection result of the ongoing process. 

\section*{Acknowledgments}
We would like to thank Eamonn Keogh and all the other people who have contributed to the UCR Time Series Data Mining Archive. This work was supported by JSPS KAKENHI Grant Number 19H04129.

\bibliographystyle{named}
\bibliography{references}

\newpage

\vfill

\end{document}